\documentclass[journal]{IEEEtran}
\usepackage[utf8]{inputenc}
\usepackage{graphicx}
\usepackage{subfigure}
\usepackage{hyperref}
\hypersetup{colorlinks,linkcolor=blue,anchorcolor=blue,citecolor=blue}
\usepackage{multirow}
\usepackage{array}
\usepackage{booktabs}
\usepackage{adjustbox}
\usepackage{tabularx}
\usepackage{wrapfig}
\usepackage{amsmath, calc}
\usepackage{amssymb}
\usepackage{amsmath}
\usepackage{mathtools}
\usepackage{natbib}
\setcitestyle{author,year}
\usepackage{amsfonts}
\usepackage{nicefrac}       
\usepackage{microtype}      
\usepackage{color}
\usepackage{graphicx}

\usepackage{xcolor}
\usepackage{enumitem}
\usepackage{bbm}
\usepackage{dsfont}

\usepackage{xcolor,colortbl}
\usepackage{color,soul}
\usepackage{floatrow}
\usepackage[ruled,norelsize]{algorithm2e}
\usepackage{algpseudocode}
\usepackage{algorithmicx}

\title{Vision-Language Intelligence: Tasks, Representation Learning, and Large Models}
\author{Feng Li$^{1,2*\dag}$\thanks{$*$Equal contribution.} \thanks{$\dag$This work was done when Feng Li, Hao Zhang, and Shilong Liu were interns at IDEA. }, ~Hao Zhang$^{1,2*\dag}$, Yi-Fan Zhang$^3$, Shilong Liu$^{2,4\dag}$ , Jian Guo$^{2}$, Lionel M. Ni$^{1,6}$,
PengChuan Zhang$^5$, ~Lei Zhang$^{2\ddag}$\thanks{$\ddag$Corresponding author.} \\
$^1$The Hong Kong University of Science and Technology \\
$^2$International Digital Economy Academy (IDEA) \\
$^3$Chinese Academy of Science $^4$Tsinghua University $^5$Microsoft Research\\
$^6$The Hong Kong University of Science and Technology (Guangzhou)\\

}

\begin{document}

\maketitle
\begin{abstract} This paper presents a comprehensive survey of vision-language (VL) intelligence from the perspective of time. This survey is inspired by the remarkable progress in both computer vision and natural language processing, and  recent trends shifting from single modality processing to multiple modality comprehension. We summarize the development in this field into three time periods, namely task-specific methods, vision-language pre-training (VLP) methods, and larger models empowered by large-scale weakly-labeled data. We first take some common VL tasks as examples to introduce the development of task-specific methods. Then we focus on VLP methods and comprehensively review key components of the model structures and training methods. After that, we show how recent work utilizes large-scale raw image-text data to learn language-aligned visual representations that generalize better on zero or few shot learning tasks. Finally, we discuss some potential future trends towards modality cooperation, unified representation, and knowledge incorporation. We believe that this review will be of help for researchers and practitioners of AI and ML, especially those interested in computer vision and natural language processing.  
\end{abstract}

\section{Introduction}
Computer vision  (CV) and natural language processing  (NLP) are sub-fields of artificial intelligence  (AI) that focus on the simulation of human intelligence in vision and language. In the last decade, deep learning has greatly advanced single-modality learning in the two fields and led to state-of-the-art results on a series of tasks. At the core of the remarkable progress of deep learning lies the empowerment of rapidly evolving GPUs and the availability of large-scale datasets, which allows for accelerated training of deep models at scale.

Along with the advancement in deep learning, we have witnessed a series of powerful neural networks developed. Traditional neural networks are typically multi-layer perceptron (MLP) consisting of multiple stacked linear layers and non-linear activations~\citep{rosenblatt1957perceptron, rosenblatt1961principles}. ~\citet{lecun1998gradient} proposed Convolutional Neural Network (CNN) to incorporate the shift-invariant property as a better inductive bias for 2D visual input, which inspired a large number of deep neural networks, including AlexNet~\citep{krizhevsky2012imagenet}, VGGNet~\citep{simonyan2014very}, GoogleNet~\citep{szegedy2015going}, and ResNet~\citep{he2016deep}. Another prominent breakthrough is recurrent neural network (RNN) in the field of natural language processing (NLP), which proposed recurrent cells for sequential data modeling~\citep{rumelhart1985learning, hochreiter1997long}. To mitigate the vanishing and exploding gradient issues in long sequence training,  LSTM~\citep{hochreiter1997long}, a variant of RNN, and GRU~\citep{chung2014empirical}, a more efficient version of LSTM, were proposed accordingly. Another great breakthrough in NLP is Transformer~\citep{vaswani2017attention}, which utilizes the attention mechanism to pursue better language representation. Using multiple stacked attention layers, Transformers can fuse information over language tokens globally with high parallelism, which facilitates both powerful representation and large-scale training.

Though inspiring progress has been achieved in single-modality domains, real-world problems often involve multiple modalities. For example, an autonomous vehicle should be able to process human orders (language), traffic signals (vision), and road conditions (vision and sounds). Even single modality learning benefits from multi-modality. For example, language learning needs perception which forms the basis of many semantic axioms~\citep{bisk2020experience}. Perception 
is the way humans understand the physical world and decides the assumption behind the human language. Since we all hear and see the same thing, we will leave some knowledge as common sense which is unwritten in our language~\citep{bisk2020experience}. Even restricted to language, speech contains more useful information than text-only, e.g., emotions can be implied through prosody. Noticing that multi-modal perception helps in both multi-modal and single-modal tasks, there comes a lot of research works. Within the field of multi-modality, the integration of vision and language gets much attention since vision is one of the most important perceptions for the human to understand the environment and language-aligned visual features can greatly improve the performances of both vision tasks and vision-language tasks. Moreover, the popularity of vision-language intelligence is also due to the availability of abundant datasets and benchmarks in this field.

The ambition to address many task-specific VL problems has fueled the initial development of VL learning. These VL problems include image captioning, visual question answering (VQA), image-text matching, etc. \citet{xu2015show, karpathy2014deep, vinyals2015show} integrated a CNN image encoder and an RNN text decoder for image captioning. \citet{antol2015vqa, yang2016stacked, anderson2018bottom} addressed the VQA task by mapping images and texts into the same latent space and predicting answers from the latent representations. \citet{kiros2014unifying, karpathy2014deep, huang2016instanceaware, lee2018stacked} performed image-text matching by calculating the similarity between an image and a text either on sentence-level or token-level. These models are tailored for specific problems with various datasets and can only solve one task each. \par

Inspired by the prevalence of pre-training and fine-tuning in both language~\citep{devlin2018bert} and vision, the interdisciplinary field of vision and language embraces a new era: to learn a joint representation of vision and language by pre-training on image-text pairs. The surge of VLP models is mostly inspired by language models in both architecture design and training methods. For example, many recent  studies~\citep{li2019visualbert, lu2019vilbert, zhang2021vinvl,tan2019lxmert,li2020oscar,yu2020ernie,chen2020uniter} adopted BERT-like~\citep{devlin2018bert} architectures and training methods. The development of VL learning meets a serious challenge due to the lack of sufficiently large scale manually labeled data. Recently, some studies~\citep{radford2021learning, jia2021scaling,wang2021simvlm,li2021grounded} broke this limitation by adopting contrastive learning and making use of large-scale web-crawled data to learn visual-linguistic features which can be used for zero-shot learning.
\par

The fast-evolving progress in the VL space urges a comprehensive survey of existing studies in this domain. This paper aims to provide a structured review of recent progress in the VL domain to help researchers gain a whole picture and better insight behind recent studies. We divide the development of VL learning into three eras. The first is from $2014$ to $2018$ where specialized models are designed for different tasks. The second era is from $2019$ to $2021$, during which joint representations of vision and language are learned by pre-training on well-labeled VL datasets. Finally, the third era began in $2021$ with the appearance of CLIP \citep{shen2021clip}, in which researchers seek to pre-train VL models on larger weakly-labeled datasets and to obtain a strong zero/few-shot vision model with VL pre-training.

When reviewing the whole development of VL intelligence, we find the general goal is to learn good visual representations. A good visual representation should have three attributes as summarized in \citep{li2021grounded}, which are object-level, language-aligned, and semantic-rich. Object-level means the granularity of vision and language features should be as fine as in object and word-level, respectively. Language-aligned emphasizes that the vision feature aligned with language can help in vision tasks. Semantic-rich means the representation should be learned from large-scale data without domain restriction. Research works in the first era of VL aim to solve specific problems instead of learning the aforementioned good representations. In the second era, researchers train models on image-text pairs to obtain language-aligned visual features. Some works in this era adopt detected regions for image representation to learn object-level features. Only in the third era can researchers deal with large-scale datasets and pre-train semantic-rich features.

There are also other survey paper discussing VL intelligence. \citet{zhang2020multimodal} analyzed multi-modal deep learning from three perspectives: multimodal representation, fusion multimodal signals and multimodal applications. \citet{mogadala2021trends} organized their survey by tasks. They reviewed some common tasks and corresponding methods. They also included  datasets and metrics. Different from them, we view the VL intelligence from the perspective time. Further more, we include the general goal of this area and show how research works atain the goal step-by-step. 

To the best of our knowledge, this is the first VL survey that summarizes studies from the viewpoint of the time period. The remainder of this paper is organized as follows. We start with some task-specific problems in VL such as image captioning, VQA, and image-text retrieval in Section II. Then we comprehensively explain the vision-language joint representation learning empowered by pre-training in Section III. Finally, we show some work learning language-aligned visual representations directly from raw image-text data with large-scale vision-language pre-training in Section VI. 

\section{Task specific problems}
\begin{figure}[h]
    \includegraphics[width=\textwidth]{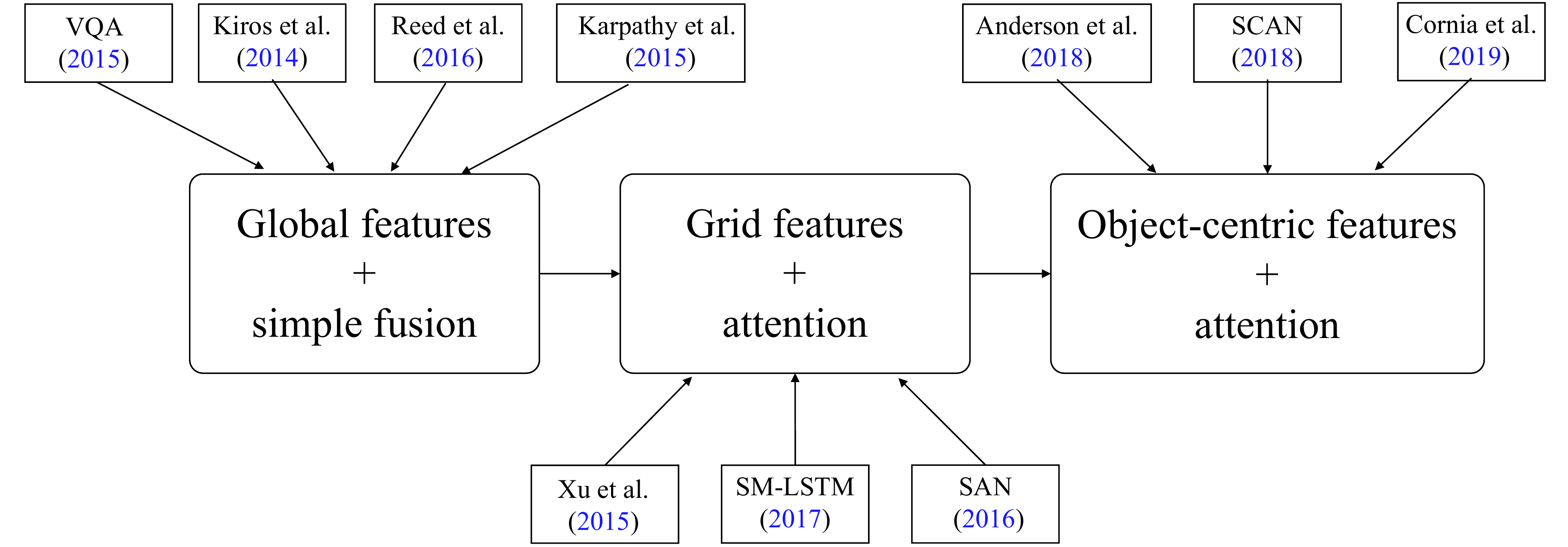}
    \caption{The three stages of task specific methods. The main differences are the granularity of the visual representation and the way of fusing vision and language features.}
    \label{fig:task specific trend}
\end{figure}

\begin{table*}
	\scriptsize
	\centering
	\caption{A comparison over task-specific problems. Tasks are classified into four categories. For each task, we have summarized the input, output, datasets, metrics, and mainstream methods.}
	\label{table:cmparison}
	\begin{tabular}{|c|c|c|c|l|l|l|}
		\hline
		&Tasks & Input &Output
		& Datasets &  Metrics & Mainstream methods
		\\
        \hline
		\multirow{2}*{\begin{tabular}[c]{@{}c@{}}Generation\end{tabular}}&\begin{tabular}[c]{@{}c@{}}Image\\ Captioning\\(IC)\end{tabular} & Image  & Sentence  & \begin{tabular}[c]{@{}l@{}}COCO~\citeyearpar{lin2014microsoft},\\ Flickr30K~\citeyearpar{young2014image},\\ Flickr8K~\citeyearpar{hodosh2013framing},\\ CC3M~\citeyearpar{sharma2018conceptual},\\CC12M~\citeyearpar{changpinyo2021conceptual},\\ SBU Captions~\citeyearpar{ordonez2011im2text}
		
		\end{tabular}&\begin{tabular}[c]{@{}l@{}}BLEU~\citeyearpar{papineni2002bleu},\\ METEOR~\citeyearpar{banerjee2005meteor},\\ROUGE~\citeyearpar{lin2004rouge},\\CIDEr~\citeyearpar{vedantam2015cider},\\ SPICE~\citeyearpar{spice2016}\end{tabular}
		& \begin{tabular}[c]{@{}l@{}}Show and Tell~\citeyearpar{vinyals2015show},\\ Karpathy et al.~\citeyearpar{karpathy2015deep},\\Xu et al.~\citeyearpar{xu2015show},\\m-RNN~\citeyearpar{mao2015deep},\\
		BUTD~\citeyearpar{anderson2018bottom},\\Lu et al.~\citeyearpar{lu2017knowing},\\AoANet~\citeyearpar{huang2019attention},\\ Lu et al.~\citeyearpar{lu2018neural},\\ Cornia et al.~\citeyearpar{cornia2019show}\\ AutoCaption~\citeyearpar{zhu2020autocaption},\\ ORT~\citeyearpar{herdade2019image},\\ CPTR~\citeyearpar{liu2021cptr} \end{tabular}\\
		
		\cline{2-7}
		&\begin{tabular}[c]{@{}c@{}}Text-to-Image\\ Generation\end{tabular}& Text&Image   &\begin{tabular}[l]{@{}l@{}}COCO~\citeyearpar{lin2014microsoft},\\ CUB~\citeyearpar{wah2011caltech} \end{tabular} &\begin{tabular}[c]{@{}l@{}}Inception score~\citeyearpar{salimans2016improved}\\ FID~\citeyearpar{heusel2017gans}\\R-precision\end{tabular}  &   
		\begin{tabular}[l]{@{}l@{}}StackGAN~\citeyearpar{zhang2017stackgan},\\AttnGAN~\citeyearpar{xu2018attngan},\\ ObjGAN~\citeyearpar{li2019object} \end{tabular}
		\\
		
		\hline
		
		\multirow{4}*{
		\begin{tabular}[c]{@{}c@{}}\end{tabular}}&\begin{tabular}[c]{@{}c@{}}Visual Question\\and Answer\\
		(VQA)\end{tabular} & \begin{tabular}[c]{@{}c@{}}Image+\\Text\end{tabular}  & Phrase  &\begin{tabular}[c]{@{}l@{}}VQA~\citeyearpar{antol2015vqa},\\ VQA v2~\citeyearpar{balanced_vqa_v2},\\ DAQUAR~\citeyearpar{malinowski2014multi},\\ Visual Genome~\citeyearpar{krishna2017visual},\\ COCO QA~\citeyearpar{ren2015exploring}
		 \end{tabular} 
		 &\begin{tabular}[c]{@{}c@{}}VQA Accuracy ~\citeyearpar{antol2015vqa}\end{tabular}  
		 &	\begin{tabular}[c]{@{}l@{}}Antol et al.~\citeyearpar{antol2015vqa},\\
		 Kim et al.~\citeyearpar{kim2016multimodal},\\
		 Ren et al.~\citeyearpar{ren2015exploring},\\
		 Malinowski et al.~\citeyearpar{malinowski2015ask},\\
		 Gao et al.~\citeyearpar{gao2015talking},\\
		  SAN~\citeyearpar{yang2016stacked},\\
		  BUTD~\citeyearpar{Anderson_2018_CVPR},\\
		 MCB~\citeyearpar{fukui2016multimodal},\\
		 MUTAN~\citeyearpar{ben2017mutan}
		 \end{tabular}\\ 
		\cline{2-7}
		Understanding&\begin{tabular}[c]{@{}c@{}}Visual Dialog\\(VisDial)\end{tabular} & \begin{tabular}[c]{@{}c@{}}Image+\\Dialog+\\Sentence\end{tabular} & Sentence  & \begin{tabular}[c]{@{}l@{}}VisDial~\citeyearpar{das2017visual},\\ GuessWhat?!~\citeyearpar{de2017guesswhat}, \\Image Chat~\citeyearpar{shuster2018image},\\ CLEVER~\citeyearpar{kottur2019clevr}\end{tabular} & Mean Rank, MRR, Recall & VisDial~\citeyearpar{das2017visual}\\
		\cline{2-7}
		
		&\begin{tabular}[c]{@{}c@{}}Visual\\ Reasoning\\(VR) \end{tabular}&\begin{tabular}[c]{@{}c@{}} Image+\\Text+\\Graph\end{tabular}&Text   &\begin{tabular}[c]{@{}l@{}} GQA~\citeyearpar{hudson2019gqa},\\ CLEVER~\citeyearpar{kottur2019clevr},\\NLVR~\citeyearpar{suhr2017corpus},\\NLVR$^2$ \citeyearpar{suhr2019corpus},\\ VCR~\citeyearpar{zellers2019recognition} \end{tabular} &Accuracy  &   
		\begin{tabular}[l]{@{}l@{}}
		NMN~\citeyearpar{andreas2017neural},\\N2NMN~\citeyearpar{hu2017learning},\\PG+EE~\citeyearpar{johnson2017inferring},\\TbD-net~\citeyearpar{mascharka2018transparency},\\StackNMN~\citeyearpar{hu2019explainable},\\NS-VQA~\citeyearpar{yi2019neuralsymbolic},\\XNM-Det~\citeyearpar{shi2019explainable}
		\end{tabular}
		\\
		\cline{2-7}
		&{\begin{tabular}[c]{@{}c@{}}Visual\\ Entailment\\ (VE)\end{tabular}}&
		{\begin{tabular}[c]{@{}c@{}}Image+\\Text\end{tabular}}
		&label&SNLI-VE~\citeyearpar{xie2019visual}&Accuracy&
	
EVE-ROI~\citeyearpar{xie2019visual} \\
		\hline
		\multirow{2}*{\begin{tabular}[c]{@{}c@{}}Retrieval\end{tabular}}&\begin{tabular}[c]{@{}c@{}}Image-text\\ Retrieval \\(IR)/\\  Text-image\\ Retrieval\\(TR)\end{tabular}& Text/Image&Image/Text   &\begin{tabular}[l]{@{}l@{}}COCO~\citeyearpar{lin2014microsoft},\\ Flickr30K~\citeyearpar{young2014image},\\ Flickr8K~\citeyearpar{hodosh2013framing}  \end{tabular} &Recall@K, Median r  &   
		\begin{tabular}[l]{@{}l@{}}
		Frome et al.~\citeyearpar{41869},\\
		Socher et al.~\citeyearpar{socher-etal-2014-grounded},\\
		
		Deep Fragment~\citeyearpar{karpathy2014deep},\\ MNLM~\citeyearpar{kiros2014unifying},\\
		Xu et al.~\citeyearpar{xu2015show},\\
		m-CNN~\citeyearpar{ma2015multimodal},\\ Klein et al.~\citeyearpar{klein2015associating},\\ Wang et al.~\citeyearpar{wang2016learning},\\ m-RNN~\citeyearpar{mao2015deep},\\ Show and Tell~\citeyearpar{vinyals2015show},\\ Huang et al.~\citeyearpar{huang2016instanceaware}\\
		Nam et al.~\citeyearpar{nam2017dual},\\
		SCAN~\citeyearpar{lee2018stacked},\\
		Faghri et al.~\citeyearpar{faghri2018vse} \end{tabular}
		\\
		
		\hline
		\multirow{2}*{\begin{tabular}[c]{@{}c@{}}Grounding\end{tabular}}
		&{\begin{tabular}[c]{@{}c@{}}Phrase\\ Grounding\\(PG)\end{tabular}}
		&\begin{tabular}[c]{@{}c@{}}Image+\\ Phrases\end{tabular} &\begin{tabular}[c]{@{}c@{}}Bounding\\ Boxes  \end{tabular}
		&\begin{tabular}[c]{@{}l@{}}
		Visual Genome~\citeyearpar{krishnavisualgenome},\\Flicker30K~\citeyearpar{young2014image},\\Flicker30K Entities~\citeyearpar{plummer2016flickr30k}
		\end{tabular} 
		&\begin{tabular}[c]{@{}l@{}}Phrase localization Accuracy,\\Recall@K\end{tabular}  
		&\begin{tabular}[l]{@{}l@{}}
		MDETR~\citeyearpar{kamath2021mdetr},\\Gupta et al.~\citeyearpar{gupta2020contrastive}\\Align2Ground~\citeyearpar{datta2019align2ground}\\
		\end{tabular}
		\\
		\cline{2-7}
		&{\begin{tabular}[c]{@{}c@{}}Reference Expression\\ Comprehension\\(RE)\end{tabular}}&
		\begin{tabular}[c]{@{}c@{}}
		Image+\\ Text\end{tabular}
		&
		\begin{tabular}[c]{@{}c@{}}
		Bounding\\ Boxes\end{tabular}
		&\begin{tabular}[c]{@{}l@{}}
		RefCOCO(+/g)~\citeyearpar{yu2016modeling,kazemzadeh2014referitgame,mao2016generation},\\ Talk2Car~\citeyearpar{Deruyttere2019Talk2Car}\\
		Visual7W~\citeyearpar{zhu2016visual7w}
		\end{tabular}
		&Accuracy&
		\begin{tabular}[c]{@{}l@{}}
		MDETR~\citeyearpar{kamath2021mdetr},\\
		MAttNet~\citeyearpar{yu2018mattnet},\\
		MAC~\citeyearpar{hudson2018compositional}
		\end{tabular}
		\\
		\hline

	\end{tabular}
	\label{task summary}
\end{table*}
Early VL methods are designed for specific tasks. The VL domain encompasses a broad range of tasks, including image captioning, VQA, image-text matching, visual grounding, and visual dialog, etc.  Some common VL tasks are summarized in Table \ref{task summary}, which shows the input, output, datasets, metrics, and mainstream methods of each task. In this section, we only introduce the three most common tasks in detail, including image captioning, VQA, and image-text matching. For these three tasks, we will introduce their task formulation and the development of mainstream methods. For the remaining tasks, a short description of each task is included. We summarize that the development of task-specific methods is from global representations to fine-grained object-centric representations. Most VL tasks experience three stages. The first stage is \textbf{gloabl vector representation and simple fusion}. The second stage is \textbf{grid feature representation and cross-modal attention}. The third stage is \textbf{object-centric feature representation and bottom-up top-down attention~\citep{anderson2018bottom}}. The three stages and representative work are shown in Figure \ref{fig:task specific trend}.
\subsection{Image Captioning}
\textbf{Task definition:} The goal of image captioning is to generate a caption for a given image. A caption is a sentence to summarize the content of the image.
The caption usually contains objects of interest, what they are doing, and the positional relations among them.

\textbf{Methods:} Early image captioning methods before deep learning \citep{kulkarni2013babytalk,farhadi2010every} are mainly rule-based. They first recognize objects and their relations and then generate captions based on predefined rules. Early methods have limited impact since their visual recognizers have limited vocabulary and the rule-based method can not deal with complex scenarios in human language.

\begin{figure}[h]
    \includegraphics[width=\textwidth]{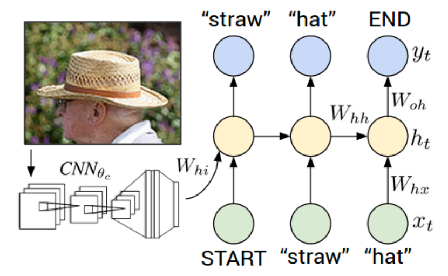}
    \caption{The img2seq structure contains an image encoder such as a CNN and a language decoder such as an LSTM.}
    \label{fig:img2seq}
\end{figure}

The breakthrough in deep learning has greatly empowered image captioning. Seq2Seq~\citep{sutskever2014sequence} achieved great success in machine translation by utilizing a text encoder to encode text from the source language and a text decoder to generate text from the target language. Following the encoder-decoder structure in Seq2Seq, \citet{xu2015show} proposed to replace the text encoder with an image encoder using GoogleNet \citep{szegedy2014going} and achieved state-of-the-art performance at that time. The encoder-decoder structure became popular and was widely adopted by later works. This structure is called img2seq as shown in Fig.\ref{fig:img2seq}.
Early studies \citep{xu2015show,karpathy2015deep, vinyals2015show} adopted a CNN model as the image encoder to extract a global CNN feature, which is fed into the text decoder as the initial hidden state. m-RNN \citep{mao2015deep} and LRCN \citep{donahue2015long} proposed to feed global CNN feature to each step of the  LSTM decoder. 

Global CNN feature has an obvious weakness because the decoder cannot focus on important regions of an image as humans do. To address this problem, attention mechanisms was introduced. \citet{xu2015show} proposed the first method leveraging attention over gird features. Assume the output feature map of a CNN feature extractor has a shape $(H, W, C)$ where $H, W$ are the height and width of the feature map and $C$ is the feature dimension. The feature map can be flattened along spatial dimensions to $H\times W$ grid features with dimension $C$. For each cell of the LSTM decoder, the hidden state attends to grid features to decide which grids to focus on. 
 
Compared to convolution, the attention mechanism has the following benefits. It allows the model to focus on certain parts of an image by giving high attention weights to important grid features. Moreover, the model learns alignments that correspond strongly with human intuition. The explainability of the model can also be improved through visualization of the attention scores, which potentially helps to troubleshoot.

However, splitting an image into equally-sized grids is a naive way to perform attention because grids align poorly with objects. To address this problem, some researchers seek to align attention with more meaningful regions. \citet{Anderson_2018_CVPR} proposed a bottom-up and top-down attention (BUTD) approach to aligning attention with salient regions that are acquired with a detection model. BUTD extracts region features with a Faster-RCNN \citep{ren2016faster} model pre-trained on Visual Genome~\citep{krishnavisualgenome}. Because detected object regions normally contain meaningful visual concepts and align better with human language, BUTD significantly improves the performance of both image captioning and VQA. Therefore, the pre-trained detector is widely adopted by subsequent VL studies.

There are also some variants of how to attend. For example, \citet{lu2017knowing} claim that the decoder does not need to always attend to visual features, since some words are irrelevant to visual features. Therefore, they proposed to use a gate to decide whether to attend or not. AoA \citep{huang2019attention} designed a special "Attention on attention" for image captioning task. After standard attention, they concatenate the attended vector and the query. Then they generate an
information vector and an attention gate from the concatenated vector and multiply the gate and information vector to obtain the output.

Except for works with attention mechanism, there are also works that do not use attention. For example, Neural Baby Talk~\citep{lu2018neural} first generates a sentence template and then fill it with concepts detected in the image. \citet{cornia2019show} generate a sentence by predicting a sequence of noun chunks. They first detect regions and then sort the regions with a sorting network. Finally, each region will be converted to a noun chunk to form the sentence.

In summary, there are two main aspects in the development of early image captioning methods, i.e. visual representation and language decoding. Visual representation develops from image-level global features to fine-grained and object-level region features and language decoding develops from LSTM to attention-based models.

\subsection{VQA}
\textbf{Task definition:} Given an image-question pair, VQA requires answering a question based on the image. Most studies treat VQA as a classification problem on a predefined answer set. For example, VQA v2 \citep{balanced_vqa_v2} has around $2$K predefined answers.

\begin{figure}[h]
    \includegraphics[width=\textwidth]{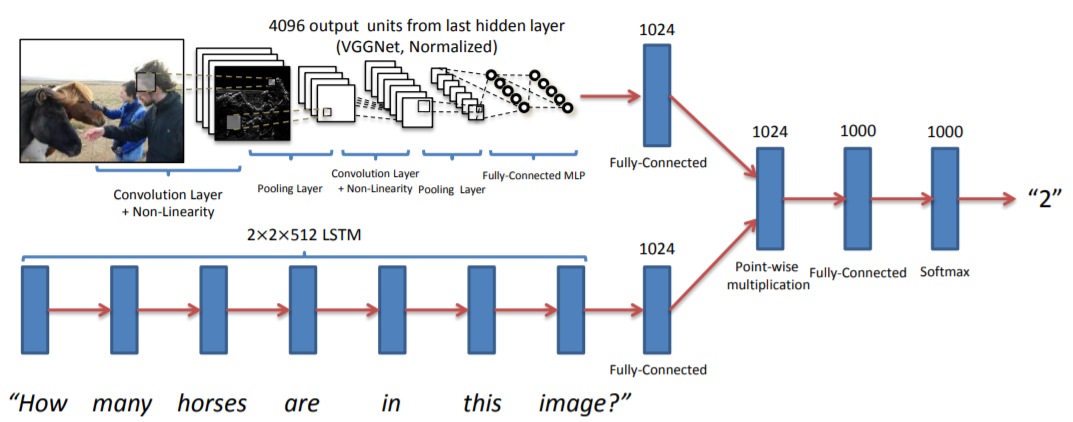}
    \caption{The architecture of vanilla VQA~\citep{antol2015vqa} contains a CNN model to encode the input images and an LSTM model to encode the input question. The encoded image and question features are merged with dot product and then go through a fully connected layer to predict the probability over candidate answers.}
    \label{fig:VQA arch}
\end{figure}

\textbf{Methods:}
The vanilla VQA \citep{antol2015vqa} is a combination of an LSTM~\citep{10.1162/neco.1997.9.8.1735} question encoder and a VGG~\citep{simonyan2015deep} image encoder. The output image embedding and question embedding are simply fused by point-wise multiplication. Then the fused vector goes through a linear layer followed by a softmax layer to output the probability of choosing each candidate answer. The architecture of the model is shown in Figure \ref{fig:VQA arch}.
Follow-up studies in VQA usually adopt the same prototype. 

Early studies~\citep{antol2015vqa,malinowski2015ask,gao2015talking} normally adopted global image representation and simple fusion. \citet{malinowski2015ask} proposed to feed the CNN image feature into each LSTM cell of the question encoder. \citet{gao2015talking} used a shared LSTM to encode the question and decode the answer. They fuse the CNN image feature with the output of each decoder cell and generate the answer word by word.

Question answering is usually only related to some regions of the image. Therefore, global representation only leads to a sub-optimal solution due to the noise introduced by irrelevant regions. \citep{yang2016stacked} proposed Stacked Attention Network (SAN) to stack multiple question-guided attention layers. In each layer, the semantic representation of the question is used as a query to attend to image grids. SAN is the first work that verifies the effectiveness of attention in VQA. \citet{fukui2016multimodal} also adopted grid features, while they fuse image and language features through bilinear pooling~\citep{lin2017bilinear}.

Grid features have limited power as we have illustrated in the image captioning task. \citet{shih2016look} proposed to use region features as visual representations. They adopted Edge Boxes~\citep{zitnick2014edge} to locate regions. BUTD \citep{anderson2018bottom} pre-trained a powerful detector and uses the question features as queries to attend to region features.  \citet{lu2017hierarchical} argued that attention on text is of equal importance as on the image. Therefore, they developed a co-attention method that jointly performs text-guided image attention and image-guided text attention.

Except for attention, there are other strategies for modality fusion. \citet{ren2015exploring} treat an image feature as a language token. They concatenates the image embeddings with language tokens as the input to LSTM. ~\citet{kim2016multimodal} proposed an iterative way of element-wise multiplication for modality fusion named Multimodal Residual Networks (MRN). MUTAN~\citep{ben2017mutan} presented parameterized bilinear interactions between modalities. Although there are many ways to fuse image and language features, attention is the most widely used one.

The core of VQA is to obtain a joint representation of image and language (the question). Researchers in this field pursued various ways to better encode and fuse image and language,  which have laid the foundation for the following VLP methods. Most works in this field encode image and language independently and then fuse them, which is similar to dual-stream methods for VLP. \citet{ren2015exploring} treat image embedding as a language token, which is similar to single-stream methods.

\subsection{Image Text Matching}
\textbf{Task definition:} Image-text matching (ITM), or say image-text retrieval, is one of the fundamental topics in vision. Given a query in a certain modality  (vision or language), it aims to find the semantically closest target from another modality. Depending on the query and the target modality, it contains two sub-tasks: image-to-text retrieval and text-to-image retrieval.\par
\textbf{Methods:} The core of image-text matching is to calculate the similarity or distance between an image and a piece of text. A widely adopted prototype is to map image and text into a shared embedding space and then calculate their similarity. The matched image and sentence are expected to have the highest similarity. 

Early methods \citep{41869,socher-etal-2014-grounded,kiros2014unifying} mainly adopted global feature to encode image and text. \citet{kiros2014unifying} proposed to learn cross-view representation with a hinge-based triplet ranking loss. \citet{faghri2018vse} considered hard negatives to improve the performance. 

\begin{figure}[h]
    \includegraphics[width=\textwidth]{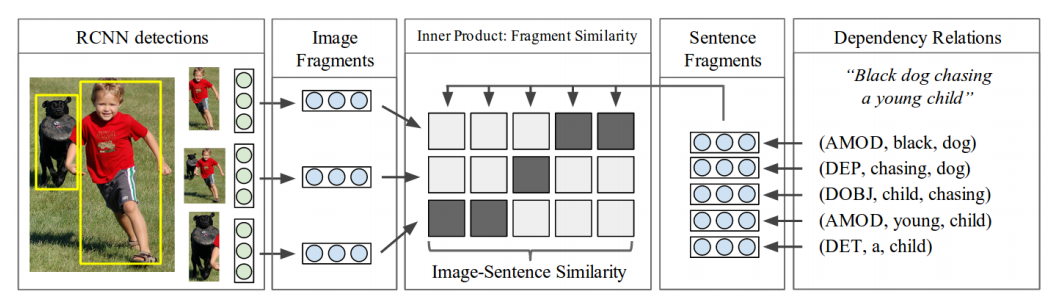}
    \caption{The overview of the Deep fragment\citep{karpathy2014deep} architecture. \textbf{Left:} Detected objects are mapped to fragment embedding space. \textbf{Right:} Dependence tree relations are encoded to fragment embedding space. }
    \label{fig:align}
\end{figure}

 \citet{karpathy2014deep} proposed Deep Fragment which is the first attempt to use fine-grained representation on both the image side and text side. The architecture of Deep Fragment is shown in Fig. \ref{fig:align}. Instead of directly representing the whole image and sentence, they map each image fragment and sentence fragment into the cross-modality embedding space. Then they align fragments between different modalities. Since one image region may be related to several words, they find the most similar region embedding for each word embedding. The similarity between the image and the sentence is the sum of the similarities between aligned word and region pairs. 

As the attention mechanism has shown great success in other VL tasks,  \citet{huang2016instanceaware} proposed to introduce attention into image-text matching (ITM). They developed a context-modulated attention scheme to attend to instance pairs appearing in both image and text. \citet{nam2017dual} proposed a dual attention framework that attends to specific regions in images and words
in the text through multiple steps and gathers essential information from both modalities. These methods proved the effectiveness of attention in the ITM task. However, as a limitation, they are multi-step methods and can only focus on one semantic part at a time.

\citet{lee2018stacked} proposed a cross-attention algorithm called SCAN to calculate the similarity between image and sentence. To enable cross attention, they represent an image as a set of regions and a sentence as a set of words. The core idea of cross attention is to not only use the sentence as a query to attend to image regions but also use the image as a query to attend to words.

Essentially, image-text matching is a problem of calculating the similarity between image and text. Early works encode image and text into global features and calculate their cosine similarity through dot product. Subsequent works adopt fine-grained features---object-level features for images and word-level features for language. They also develop more complex algorithms for calculating similarities such as cross attention.
\begin{figure*}
\subfigure[]{
\begin{minipage}{0.4\textwidth}
\centering
\label{fig:bert}
\includegraphics[width=2.3in]{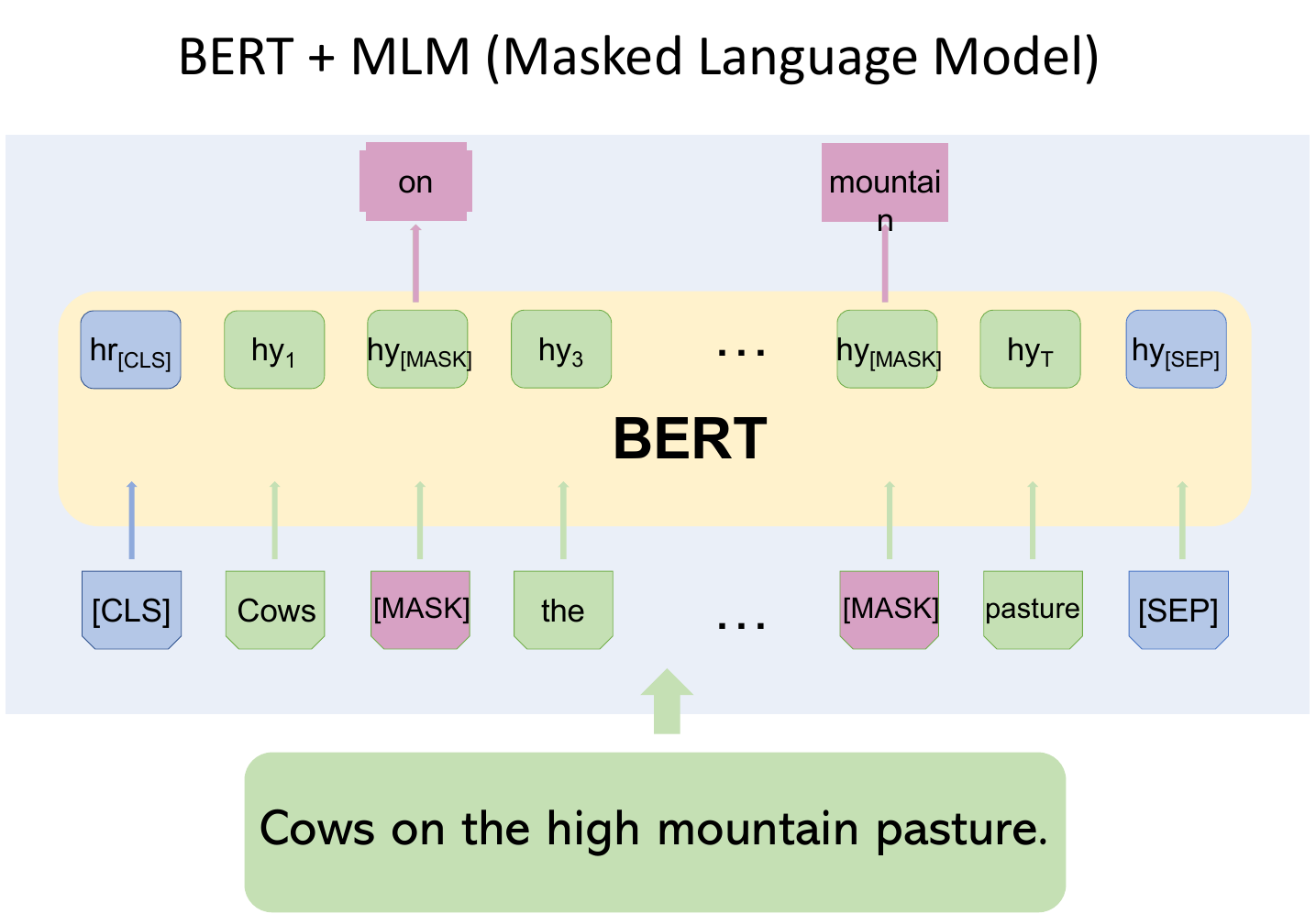}
\end{minipage}}
\subfigure[]{
\begin{minipage}{0.5\textwidth}
\centering
\label{fig:bertvl}
\includegraphics[width=3.4in]{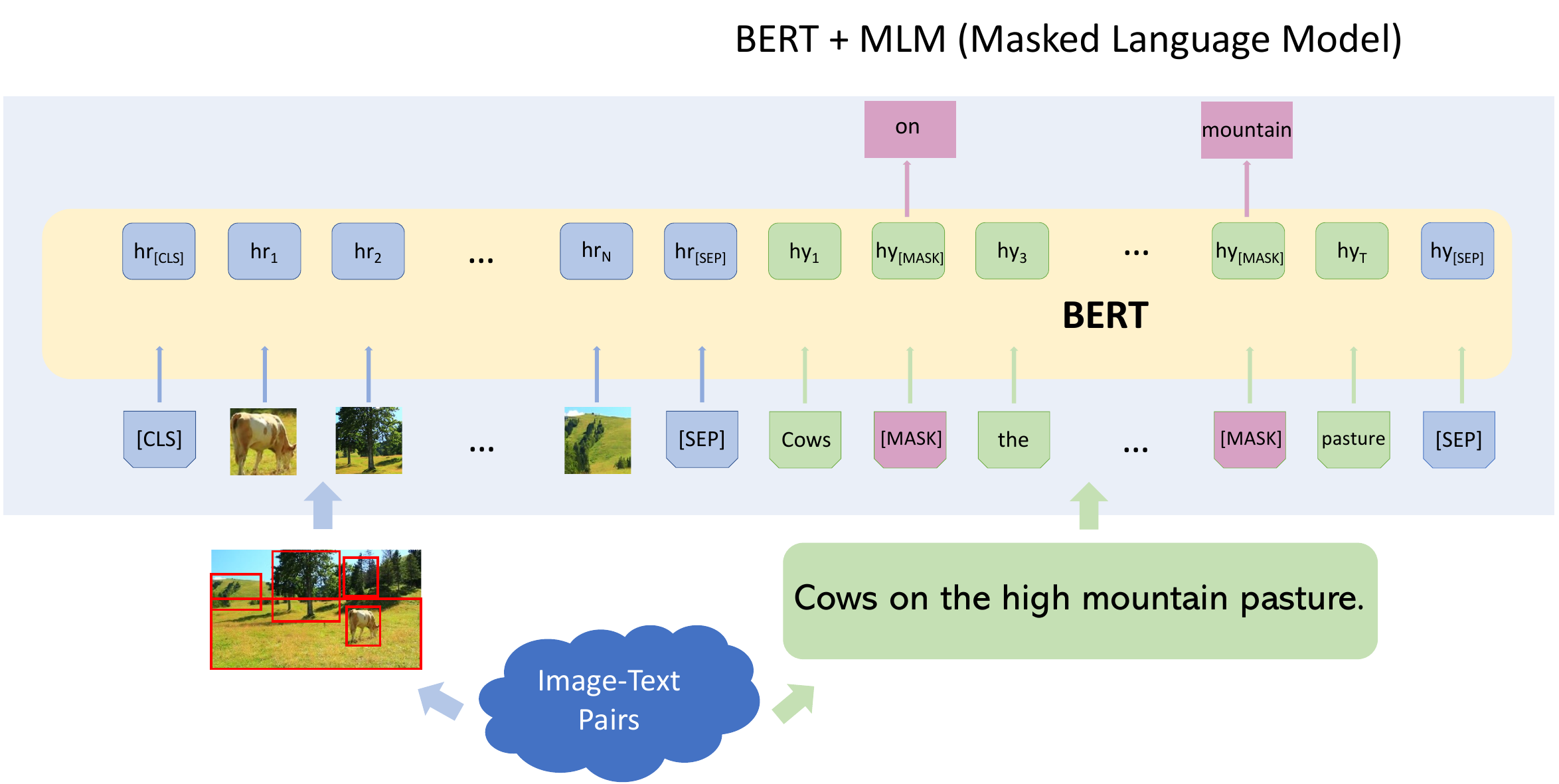}
\end{minipage}}
\caption{ (a) Original BERT with single-modality, where some language tokens are masked for prediction to train language representation.  (b) Modified BERT with multi-modality, where both image and language tokens are fed into a BERT-like Transformer model.}
\centering
\end{figure*}
 
 \subsection{Other tasks}
 There are a broad variety of tasks in the interdisciplinary field of vision and language that we can not elaborate on in detail. Therefore, we list some of the important tasks in Table~\ref{table:cmparison}, including:\par
 \textbf{Text-to-Image Generation:} Given a piece of text, generate an image containing the content of the text. We leave the details of text-to-image generation to Section \ref{t2i}.
 
 \textbf{Visual Dialog:} Given an image, a dialog history, and a question about the image, answer the question.
 
 \textbf{Visual Reasoning:} Similar to VQA, which requires answering a question about an input image, visual reasoning requires a further ability to understand the image. Visual reasoning task usually contains adequate annotations about the objects in an image, the structure of the questions, etc.
 
 \textbf{Visual Entailment:} Given an image and a text, decide whether the image semantically entails the input text.
 
 \textbf{Phrase Grounding and Reference Expression Comprehension:} These two tasks require a model to output bounding boxes corresponding to the text. For phrase grounding, the text is a set of phrases and for reference expression comprehension, the text is an expression.
 
In the era of task-specific methods, researchers design specific models for different tasks. Although the models for different tasks vary significantly, they follow a similar trajectory. They all have three stages as shown in Fig.~\ref{fig:task specific trend}. The technological development of this era laid the foundation for the VLP era.

\section{Vision Language Joint Representation}
\label{Vision Language Joint Representation}

The pre-training and fine-tuning paradigm has been adopted across a wide range of domains and advanced various downstream tasks. Among the most prominent factors that leverage the prevailing large-scale pre-training is the availability of abundant datasets along with the rapid evolution of GPUs.  Motivated by the success in single-modal language/vision pre-training, researchers started to explore the joint representation of language and vision~\citep{sun2019videobert,lu2019vilbert}, giving birth to the cross-modal VLP models. 

The recent surge of VLP models is mostly inspired by language models in both architecture design and training methods. One of the most important breakthroughs is Transformer which was developed by \citet{vaswani2017attention} for improving language representation. Using multiple stacked attention layers, Transformers can fuse information over language tokens globally with high parallelism, which facilitates both powerful representation and large-scale training. A successful application of Transformer is BERT~\citep{devlin2018bert} which leverages Transformer encoders and introduces a bidirectional masking technique that allows each language token to attend to other tokens bidirectionally. As shown in Figure~\ref{fig:bert}, training is conducted by replacing some text tokens with a special [MASK] token and predicting each [MASK] using its context information. With this technique, language representation training can be considered as a denoising process, in which an input sentence learns to reconstruct itself with some contaminated tokens. This denoising training forces the masked tokens to utilize all the unmasked information, hence resulting in contextualized representations.
The architecture design and mask training technique developed for Transformer-based language models is the main principle behind a broad variety of cross-modal developments that contributes to the recent surge of VLP models. Figure~\ref{fig:bertvl} shows a simple cross-modal BERT~\citep{anderson2018bottom}. Similar to language training, image is tokenized and embedded along with language tokens with certain techniques, which will be elaborated on later. Usually, the tokenized visual features and textual features together are fed into a Transformer encoder with masked language training to learn a joint representation.
\par
\begin{figure*}[t]
    \includegraphics[width=\textwidth]{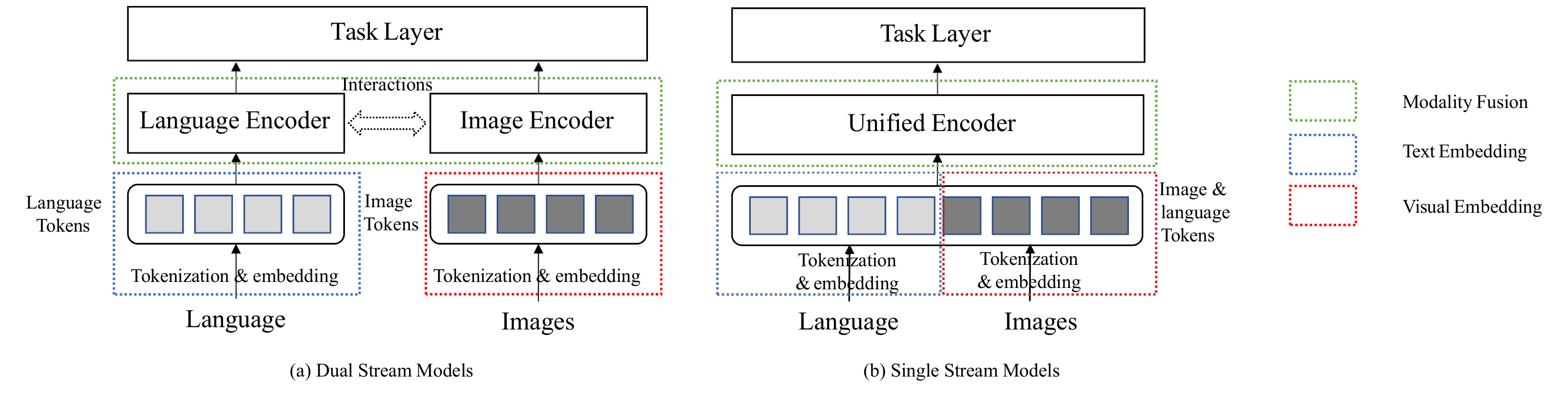}
    \caption{The architecture of VLP models usually contains Visual Embedding (VE), Textual Embedding (TE), and Modality Fusion (MF). (a) is the illustration of a dual-stream model, and (b) shows a single stream model. In a dual-stream model, modality fusion is optional and is done by the interactions (usually cross attention) between language and image encoder. In a single stream model, modality fusion is done in a unified encoder (usually multi-layer Transformers).
    }
    \label{fig:pre-train-structure}
\end{figure*}
In this section, we will go through the main components of VLP models. As shown in Figure~\ref{fig:pre-train-structure}, there are primarily three components in VLP models, namely the \textit{visual embedding (VE), textual embedding (TE), and modality fusion (MF)} modules. VE and TE are normally pre-trained with images and texts respectively, whereas MF takes the features extracted from VE and TE, and fuses them with image-text pre-training. The goal of VLP is to learn \emph{object-level, language-aligned, and semantic-rich} visual representations. Object-level means the learned representation is fine-grained and aligned with objects rather than for a whole image. Works that use features of detected objects to represent images~\citep{tan2019lxmert,lu2019vilbert,yu2020ernie,li2019visualbert,li2019unicoder,li2020oscar,hu2020vivo,li2021grounded} are object-level. Language-aligned aims to learn visual features that are well-aligned with language words, which is the goal for most VLP methods. Semantic-rich strives for a representation that can be generalized to a broad range of semantic concepts and needs to be learned from a large-scale dataset.

Pre-training on a massive dataset is crucial for improving the performance on downstream tasks with smaller datasets, as the learned representation can be transferred in downstream tasks. VLP models have been proven very effective to empower downstream tasks.

\subsection{Why Pre-training Is Needed?}
Deep learning is essentially a statistical data-driven approach, which aims to learn a mapping function from seen data so that to make predictions on unseen data using the learned mapping function. Note that the ultimate goal is to achieve a good performance on unseen data. In terms of statistical learning, such a goal is represented as minimizing an \textit{expected} loss over the whole data space which follows a fixed but unknown distribution. However, such an expected loss minimization is not tractable as the distribution is unknown. In practice, one has to sample data from this distribution and define an \textit{empirical} loss as a proxy to the expected loss. This may sound strange, but is actually a commonly used practice in machine learning. For example, for an image classification problem of telling whether an input image has a cat or not, the most practical approach is to collect training images of cat and no-cat and then train a classifier by minimizing an empirical loss defined on this training set. However, the distribution of cat and no-cat images is indeed unknown.

Statistical learning theory shows that for a sufficiently large number of independent and identically distributed (i.i.d.) data sampled from such an unknown distribution, the empirical loss minimization result converges to the expected loss minimization result. That is, asymptotically, one can use i.i.d. samples to approximate a loss function defined by an unknown distribution. However, in practice, data is never sufficient to represent the unknown distribution and thus leads to many deficiencies, such as low performance on uncovered data, being vulnerable to adversarial attacks, etc.

Pre-training allows one to leverage an unlimited amount of data without labels (or with weak labels) to learn features that conform with downstream tasks. Such a large-scale data set helps define a better approximation to the expected loss for learning more robust and non-spurious patterns from data. Thanks to the shared model between the pre-training and fine-tuning stages, with very limited (e.g., few-shot) supervision, the learned feature can lead to high accuracy on downstream tasks after fine-tuning. This makes the paradigm of pre-training and fine-tuning an effective solution to solving (or mitigating) the data shortage problem. For more discussions, see \citep{zhang2022statistical}. 

\subsection{Modality Embedding}
Text and image are different levels of information in nature concerning dimensionality and structure. To resolve this modality discrepancy, modality embedding is normally utilized to extract features from each modality independently and then map the features into a shared feature space. As shown in Figure~\ref{fig:pre-train-structure}, modality embedding involves visual embedding and textual embedding, both encompassing a tokenization process and an embedding process. Visual embedding aims to follow textual embedding to convert an image into a number of tokens with the level of representation as text tokens. 
Ablation studies conducted by ~\citet{bugliarello2021multimodal} demonstrate that training datasets and hyperparameters are responsible for most performance improvements across many different VLP models and also emphasize the importance of modality embeddings.

\subsubsection{Text Tokenization and Embedding}
\
\newline
Before text embedding, the text should be tokenized. Considering the discretization nature of language, early work simply regards each word as a token. A pioneering study is Word2Vec~\citep{mikolov2013efficient}, which proposed a continuous CBOW and a skip-gram model to train word vector representation. Word2Vec is computationally efficient to scale to large corpus and produces high-quality embeddings. However, although its vocabulary size is as large as about one million, this method suffers from out-of-vocabulary issues due to rare or unseen words, making it difficult to learn word sub-units such as 'est'. To resolve this problem, \textit{Sennrich et. al}~\citep{sennrich2015neural} proposed a subword tokenization approach, which segments words into smaller units with byte pair encoding  (BPE)~\citep{gage1994new}. Subword tokenization is widely used in many language models including BERT. 

Most VLP models adopt text embeddings from pre-trained BERT~\citep{devlin2018bert}. As BERT is trained with masked token learning using Transformer encoders, it has a strong bidirectional representation ability.

\subsubsection{Visual Tokenization and Embedding}
\
\newline
Different from language tokens that are discrete and arranged in a single dimension, images are from a high dimensional space and have co-related pixel values. Therefore, image tokenization is usually more complicated than text tokenization. Basically, image tokenization can be categorized into \emph{region based}, \emph{grid based}, and \emph{patch based}, which are introduced as follows.

\textbf{1) Grid features} are directly extracted from equally sized image grids with a convolution feature extractor as aforementioned. For example, \citet{huang2020pixel, huang2021seeing} adopted grid features as the image embedding of their VLP models. The advantages of grid features are mainly two folds. The first is convenient as it does not require a pre-trained object detector. The second is that besides salient objects, grid features also contain background which may be useful for downstream tasks.

\textbf{2) Region features} are extracted by a pre-trained object detector. Most recent VLP models adopt region features to learn object-level joint representations. Especially, most VLP models adopt Faster R-CNN trained on the Visual Genome (VG) dataset as region feature embedding following the work of BUTD~\citep{anderson2018bottom}. There are three essential components of region features, which are bounding boxes, object tags, and RoI features (feature vectors after RoI pooling). Bounding boxes are commonly used in VLP as position indicators, which are encoded through a transformation into the same dimensional space as RoI features and added to RoI features. Object tags are widely utilized in training methods such as Masked Region Classification, which will be elaborated later in \ref{section: maskVisionModelling}. The advantage of region features is that they help a VLP model focus on meaningful regions of the image. These regions are usually closely related to downstream tasks.\\

\textbf{3) Patch features} are usually extracted by a linear projection on evenly divided image patches. The main difference between patch and grid features is that grid features are extracted from the feature map of a convolutional model while patch features directly utilize a linear projection. Patch features were first introduced by Vision Transformer (ViT)~\citep{dosovitskiy2021image} and then adopted by VLP models~\citep{kim2021vilt, xue2021probing}.
The advantage of using patch features is efficiency. For example, ViLT accelerates the pre-training by $10$ times with competitive results~\citep{kim2021vilt}.

Image embedding methods normally vary for different tokenization schemes. Grid features and region features are usually from pre-trained convolutional models, whereas patch features can be simply embedded by a linear layer.

\subsection{Modality Fusion}
At the core of VLP models lies the modality fusion, which models intra-modality and inter-modality fusion to produce contextualized joint representations of image and text. MF schemas can be categorized into dual-stream modeling and single-stream modeling. The general structure of VLP is shown in Figure~\ref{fig:pre-train-structure}.

\subsubsection{Dual stream modeling}
\
\newline
Dual-stream modeling aims to map vision and language into the same semantic space. It is the seminal method for modality fusion~\citep{tan2019lxmert,yu2020ernie,li2021semvlp}. As shown in Fig.~\ref{fig:pre-train-structure}(a), it adopts two separate encoders to learn high-level representations for vision and language, respectively. The dual-stream design allows variable network depth and architecture to be adaptive to each modality. Apart from intra-modal fusion within each modality, some studies~\citep{lu2019vilbert,tan2019lxmert} also explicitly design inter-modal interactions between two encoders to enable modality fusion in different encoding stages.

\subsubsection{Single stream modeling}
\
\newline
Single stream modeling aims to learn one joint representation. Image and text tokens are concatenated and inputted into Transformers as shown in Fig.~\ref{fig:pre-train-structure}(b). Most VLP models adopt this modality fusion scheme~\citep{alberti2019fusion,chen2020uniter,li2019unicoder,huang2021seeing,jia2021scaling}. Single-stream modeling performs implicit intra-modal and inter-modal fusion, free from the architecture design of the fusion stage in dual-stream modeling.

\subsection{Training}\label{training}
To learn a joint representation of vision and language, vision language pre-training methods usually use several self-supervised learning losses to pre-train the model on a large dataset. As studied in \citep{chen2020uniter,li2019unicoder,lu2019vilbert,su2019vl,tan2019lxmert,zhou2019unified}, there are mainly three pre-training methods, which are Image Text Matching (ITM), Masked Language Modeling  (MLM), and Masked Visual Modeling (MVM).
\subsubsection{Image Text Matching}
\
\newline
The goal of ITM is to predict whether a pair of images and text is matched. ITM can be formulated as a binary classification task. Previous work \citep{chen2020uniter} applies a sigmoid function on the output of the special token $\left[CLS\right]$ to predict whether the input image and text are matched. The loss function is 
\begin{equation}
\mathcal{L}_{\text {ITM }}=-\mathbb{E}_{ (\mathcal{W}, V) \sim D} \log p (y \mid\mathcal{W}, V)
\end{equation}
where $\mathcal{W}=\{w_1, w_2, ..., w_n\}$ denotes a sequence of language tokens and $V$ denotes the visual content. $y=0$ or $1$ indicates whether the image is matched  ($y=1$) or not  ($y=0$).
\subsubsection{Masked Language Modeling}
\
\newline
According to \citep{chen2020uniter}, MLM is utilized to encourage the model to learn the implicit relation between language tokens and visual content. The goal is to reconstruct the masked language tokens from the known language tokens and visual contents. This goal can be formulated as
\begin{equation}
\mathcal{L}_{\mathrm{MLM}}=-\mathbb{E}_{ (\mathcal{W}, V) \sim D} \log p\left (w_{i} \mid \mathcal{W}_{\backslash i}, V)\right),
\end{equation}
where $\mathcal{W}_{\backslash i}$ denotes the sentence without the $i$-th word. Note that, although BPE is normally adopted for language tokenization, the minimal masked unit is a whole word instead of a subword. The reason is that a subword can be easily predicted from its surrounding subwords due to information leakage. There are also improved versions of MLM. For example, \citet{sun2019ernie} proposed Knowledge Masked Language Modeling, which performs phrase-level masking and entity-level masking to integrate phrase and entity-level knowledge into the language representation. For entity level masking, they treat a named entity as a whole. For example, J. K. Rowling, which contains three tokens, is a person name and should be masked together in entity-level masking. The phrase level masking treats a group of words as a conceptual unit. They mask all the tokens belonging to a phrase and predict them simultaneously.

\begin{figure*}[t]
    \includegraphics[width=\textwidth]{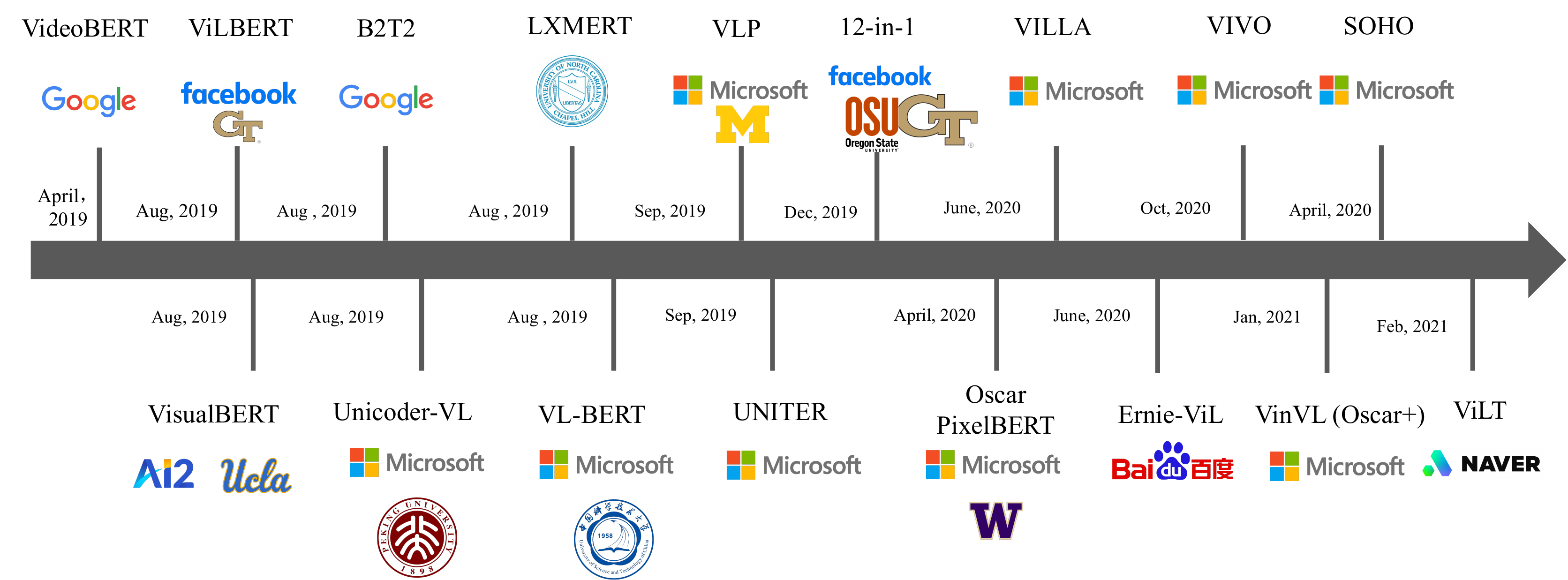}
    \caption{A Landscape of VLP methods. Works are sorted according to the time they were published. We also show the logos of main institutions where each work is from.}
    \label{fig:pre-train-roadmap}
\end{figure*}

\subsubsection{Masked Vision Modeling}\label{section: maskVisionModelling}
\
\newline
Inspired by MLM, MVM is designed to learn contextualized visual representation by reconstructing masked visual contents. MVM is more challenging than MLM since the information density of image is lower than that of language. When reconstructing a missing word, sophisticated language understanding is required. On the contrary, a missing image patch can be recovered from neighboring patches without cross-modality understanding ~\citep{he2021masked}. To overcome this gap, most works mask detected object regions that have relatively high information density. Other works such as SOHO~\citep{huang2021seeing} use a visual dictionary (VD) to represent more
comprehensive and compact semantics in the visual domain so that they can apply MVM in the same way as MLM. In summary, there are mainly four MVM schemes.
\begin{itemize}
    \item [1)] Masked Region Prediction (MRP): MRP Minimizes the distance between the predicted feature and the feature output by a pre-trained object detector. The distance metric is usually L2~\citep{tan2019lxmert}. We denote masked image regions as $\mathbf{v_m}=\left\{v_m^{1}, ..., v_m^{M}\right\}$, $h_{\theta}^i$ as the model prediction corresponding to $v_m^i$, and $r\left(v_{m}^{i}\right)$ as the ROI-pooled feature of $v_m^i$. The loss function is
    \begin{equation}\mathcal{L}_{MRP}=\sum_{i=1}^{M}\left\|h_{\theta}^i-r\left(v_{m}^{i}\right)\right\|_{2}^{2}\end{equation}
    \item [2)] Masked Region Classification (MRC): MRC requires a model to predict the object semantic class for each masked region. Since there is no ground-truth label, the label $c(v_m^i)$ predicted by a pre-trained object detector of $v_m^i$ is used as the target label of $v_m^i$. Denoting the predicted label of the VLP model as $g_{\theta}\left(v_{m}^{i}\right)$, the loss fuction is
    \begin{equation}
    \mathcal{L}_{MRC}=\sum_{i=1}^{M} \operatorname{CE}\left(c\left(v_{m}^{i}\right), g_{\theta}\left(v_{m}^{i}\right)\right)\end{equation}
    
    \item [3)] Masked Region Classification with KL-Divergence (MRC-KL): As the target label for MRC is inaccurate, MRC-KL adopts the soft label $\tilde{c}\left(v_{m}^{i}\right)$ of $v_{m}^{i}$ as the supervision signal, which is the raw output of the object detector after softmax. The loss function is
    \begin{equation}
    \mathcal{L}_{MRC-kl}=\sum_{i=1}^{M} D_{K L}\left(\tilde{c}\left(v_{m}^{i}\right) \| g_{\theta}^i\right)\end{equation}
    where $g_{\theta}^i$ denotes the soft label of $v_m^i$ predicted by the VLP model.
    \item [4)] Masked Visual Modeling with Visual Dictionary (MVMVD): Similar to language models which have a vocabulary dictionary, MVMVD requires a visual vocabulary dictionary (VD). The goal of MVMVD is to reconstruct the masked VD tokens ~\citep{huang2021seeing}.
    The loss function is 
    \begin{equation}\mathcal{L}_{\mathrm{MVM}}=-\mathbb{E}_{(\mathcal{W}, f(\mathcal{V})) \sim D} \log p\left(f\left(v_{j}\right) \mid \mathcal{W}, f(\mathcal{V})_{\backslash j}\right)\end{equation}
    where $f(\cdot)$ denotes the mapping from an image grid to a visual token in the VD and $j$ denotes the index of the masked token in the VD. 
\end{itemize}

There are two points that are worth noting. Firstly, to encourage inter-modal fusion, some works such as UNITER-VL~\citep{chen2019uniter} only mask tokens in one modality each time during training to encourage the masked tokens to attend to another modality for missing information~\citep{chen2019uniter}. Secondly, for MVMVD, neighboring image grids tend to map to the same VD token as they are highly co-related. When performing reconstruction, the model may directly copy the surrounding tokens. Therefore, all visual embedding vectors mapped to the same VD token are masked together in SOHO. Despite all the above mentioned MVM methods, effective vision modeling remains a challenging problem. The results of some ablation studies in VLP models such as SOHO~\citep{huang2021seeing} indicate that adding MVM task only yields small additional improvement to the performance. \citet{cao2020behind} reveal that VLP models exhibit the propensity to focus on textual information than visual information in downstream tasks.  

\begin{table*}
\adjustbox{max width=\textwidth}{%
	\centering
	\caption{Pre-training work comparison. The pre-training tasks correspond to the tasks described in Section \ref{training}.  We also list the pre-training datasets and downstream tasks of these works. The datasets and downstream tasks are described in Table \ref{table:cmparison}}.
	\label{table:work comparison}
	\begin{tabular}{|c|c|c|c|c|c|}
		\hline
		Architecture&Method & Visual Embedding &Pre-training Tasks
		&Pre-training Datasets &  Downstream Tasks \\
		\hline
		\multirow{3}*{\begin{tabular}[c]{@{}c@{}}\end{tabular}}
		&\begin{tabular}[l]{@{}c@{}}ViLBERT~\citeyearpar{lu2019vilbert}\end{tabular} & BUTD  & \begin{tabular}[l]{@{}c@{}}ITM, MLM\\ MRC-kl \end{tabular} & CC3M &\begin{tabular}[l]{@{}c@{}}VQA, VR, RE\\ IR, Zero-shot IR \end{tabular} \\
		\cline{2-6}
		
		&\begin{tabular}[c]{@{}c@{}}LXMERT~\citeyearpar{tan2019lxmert}\end{tabular} & BUTD  & \begin{tabular}[c]{@{}c@{}}ITM, MLM, MRP, MRC \end{tabular}  & \begin{tabular}[c]{@{}c@{}}COCO, VG~\citeyearpar{krishnavisualgenome}, VQA v2\\ GQA, Visual7W~\citeyearpar{zhu2016visual7w} \end{tabular} & VQA, VR \\
		\cline{2-6}
      Dual-Stream &\begin{tabular}[c]{@{}c@{}}12-in-1~\citeyearpar{lu202012}\end{tabular} & BUTD  & \begin{tabular}[c]{@{}c@{}}ITM,MLM,MRC-kl \end{tabular}  & \begin{tabular}[c]{@{}c@{}}VQA v2, Flickr30k\\ SNLI-VE, COCO\\ GuessWhat, VG \\ RefCOCO, RefCOCO+,RefCOCOG\\ Visual 7W, GQA, NLVR$^2$\end{tabular} & 
      \begin{tabular}[c]{@{}c@{}}VQA ,IR, RE, VE, VR\end{tabular}
      \\
        \cline{2-6}
        &\begin{tabular}[c]{@{}c@{}}Ernie-ViL~\citeyearpar{yu2020ernie}\end{tabular} & BUTD  & \begin{tabular}[c]{@{}c@{}}Object Prediction\\ Attribute Prediction \\ Relationship Prediction\\ ITM, MLM,MRC-kl \end{tabular}  &\begin{tabular}[c]{@{}c@{}}CC3M, SBU (out-of-domain)\\ COCO, VG (in-domain) \end{tabular} & \begin{tabular}[c]{@{}c@{}}VQA, VR, IT, TR, RE \end{tabular} \\
		\hline
		
        \multirow{2}*{}
      &\begin{tabular}[c]{@{}c@{}}VideoBERT~\citeyearpar{sun2019videobert}\end{tabular} & \begin{tabular}[c]{@{}c@{}}S3D~\citeyearpar{xie2017rethinking}\end{tabular}  & ITM, MLM, MVM  & YouTube cooking videos \citeyearpar{xie2017rethinking} & \begin{tabular}[c]{@{}c@{}}Zero-shot Action prediction \citeyearpar{xie2017rethinking}\\ Video Captioning \citeyearpar{xie2017rethinking}\end{tabular} \\
        \cline{2-6}
		&\begin{tabular}[c]{@{}c@{}}VisualBERT~\citeyearpar{li2019visualbert}\end{tabular}  & Pre-trained Fast R-CNN & ITM, MLM  & COCO & \begin{tabular}[c]{@{}c@{}}VQA, VR, PG\end{tabular} \\
        \cline{2-6}
		&\begin{tabular}[c]{@{}c@{}}B2T2~\citeyearpar{alberti2019fusion}\end{tabular} &  \begin{tabular}[c]{@{}l@{}}ResNet-152 \citeyearpar{he2016identity}\end{tabular} &ITM, MLM   & CC3M & VR \\
        \cline{2-6}
		&\begin{tabular}[c]{@{}c@{}}Unicoder-VL~\citeyearpar{li2019unicoder}\end{tabular} & \begin{tabular}[c]{@{}c@{}}Pre-trained\\Faster-RCNN ~\citeyearpar{singh2018pythia}\end{tabular}  & \begin{tabular}[c]{@{}l@{}}ITM, MLM, MRC \end{tabular} & CC3M, SBU & \begin{tabular}[c]{@{}c@{}}IR, TR, VR \\Zero-shot IR/TR \end{tabular} \\
        \cline{2-6}
        &\begin{tabular}[c]{@{}c@{}} VL-BERT~\citeyearpar{su2019vl}\end{tabular}& BUTD  & \begin{tabular}[c]{@{}c@{}}MLM, MRP \end{tabular}  & \begin{tabular}[c]{@{}c@{}}CC3M, English Wikipedia \\ BooksCorpus \citeyearpar{zhu2015aligning} \end{tabular} & \begin{tabular}[c]{@{}c@{}}VQA, VR, RE \end{tabular} \\
        \cline{2-6}
        
        &\begin{tabular}[c]{@{}c@{}}Unified VLP~\citeyearpar{zhou2019unified}\end{tabular} &\begin{tabular}[c]{@{}c@{}}BUTD variant\\  (with ResNext-101\\ backbone) ~\citeyearpar{xie2017aggregated}\end{tabular} &  \begin{tabular}[c]{@{}c@{}}MLM  (Sequentially\\ and bidirectionally)\end{tabular} & CC3M & \begin{tabular}[c]{@{}c@{}}IC, VQA\end{tabular} \\
        \cline{2-6}
        Single-Stream&\begin{tabular}[c]{@{}c@{}}UNITER~\citeyearpar{chen2019uniter}\end{tabular} & BUTD  & \begin{tabular}[c]{@{}c@{}}ITM, MLM, MRP, MRC\\
        MRC-kl\end{tabular}  & \begin{tabular}[c]{@{}c@{}}COCO, VG,  CC3M, SBU\end{tabular} &\begin{tabular}[c]{@{}c@{}}VQA, VR, VE, IR, TR, RE\end{tabular}  \\
        
        \cline{2-6}
        &\begin{tabular}[c]{@{}c@{}}Oscar~\citeyearpar{li2020oscar}\end{tabular} &  BUTD+tags & \begin{tabular}[c]{@{}c@{}}MLM (include tags)\\ ITM (pollute tags)\end{tabular}  & \begin{tabular}[c]{@{}c@{}}COCO, VG,  CC3M, \\SBU, fliker30k, GQA\end{tabular} &\begin{tabular}[c]{@{}c@{}}VQA, VR, VE, IR, TR\\ RE, IC\end{tabular} \\
        \cline{2-6}
        &\begin{tabular}[c]{@{}c@{}}PixelBERT~\citeyearpar{huang2020pixel}\end{tabular} &  Pixel feature embedding \citeyearpar{he2015deep} & \begin{tabular}[c]{@{}c@{}}MLM, ITM\end{tabular}  &  \begin{tabular}[c]{@{}c@{}}COCO, VG\end{tabular}&  \begin{tabular}[c]{@{}c@{}}VQA, IR, TR, VR\end{tabular}\\
        \cline{2-6}
        &\begin{tabular}[c]{@{}c@{}}VILLA~\citeyearpar{gan2020large}\end{tabular} & BUTD  & \begin{tabular}[c]{@{}c@{}}ITM, MLM, MRC-kl\end{tabular}  & \begin{tabular}[c]{@{}c@{}}COCO, VG,  CC3M, SBU\end{tabular} &\begin{tabular}[c]{@{}c@{}}VQA, VR, VE, IR, TR, RE \end{tabular}  \\
        \cline{2-6}
        
        &\begin{tabular}[c]{@{}c@{}}VIVO~\citeyearpar{hu2020vivo}\end{tabular} &  BUTD & \begin{tabular}[c]{@{}c@{}}Mask Tag Prediction\\ (Hungarian match loss)\end{tabular}  & Open Images V5~\citeyearpar{OpenImages,OpenImagesSegmentation} & IC \\
        \cline{2-6}
         &\begin{tabular}[c]{@{}c@{}}SOHO~\citeyearpar{huang2021seeing}\end{tabular} & VD  & \begin{tabular}[c]{@{}c@{}}ITM, MLM\\ MVMVD\end{tabular}  & COCO, VG & \begin{tabular}[c]{@{}c@{}}IR, TR, VQA\\ VR, VE (based on VD)\end{tabular} \\
        \cline{2-6}
		&\begin{tabular}[c]{@{}c@{}}ViLT~\citeyearpar{kim2021vilt}\end{tabular} & Patch Projection  & ITM, MLM  & COCO, VG, SBU, CC &  \begin{tabular}[c]{@{}c@{}}VQA, VR,\\IR, TR,\\Zero-shot IR\&TR\end{tabular} \\
		\hline
		Hybrid&\begin{tabular}[c]{@{}c@{}}SemVLP~\citeyearpar{li2021semvlp}\end{tabular} & Pre-tained Faster-RCNN  & MLM, MRP, VQA  & \begin{tabular}[c]{@{}c@{}}COCO, VG, VQA v2,\\ GQA, Visual7W \end{tabular} & VQA, VR, IR, TR \\
		\hline
		
	\end{tabular}}  
\end{table*}
\subsection{Landscape of General Pre-training Studies}
After introducing the general pipeline of VLP models, in this section, we summarize some pioneering works in the cross-domain of VLP.

Inspired by the success of pre-training in NLP and CV, a boosting number of research works in the domain of VLP has recently surged to pursue a unified cross-modality representation. A landscape of VLP works is shown in Figure \ref{fig:pre-train-roadmap}. A more detailed comparison of related works is shown in Table~\ref{table:work comparison}. We elaborate on some representative studies in this section.\par
\noindent\textbf{Single Stream Models: }
VideoBERT~\citep{sun2019videobert} is a pioneering work to learn joint representation for video and language. The primary idea is to feed visual and textual tokens into a single-stream model built upon BERT~\citep{devlin2018bert}. Textual tokens are extracted by converting video speech into text with an automatic speech recognition approach, and visual tokens are acquired by extracting features from video clips using a convolutional backbone. VideoBERT is capable of performing a wide range of downstream classification and generation tasks, including video captioning and zero-shot mask verbs/nouns prediction. Note that VideoBERT is pre-trained on cooking videos, where the contents are instructional and of high-quality. It assumes that spoken words are well aligned with visual content, which limits its application to only certain videos (e.g. instructional videos). Another issue confining its scalability is its delicately designed captioning text template, for example, \textit{now let’s [MASK] the [MASK] to the [MASK], and then [MASK] the [MASK]}, which only works for cooking videos.

~\citet{li2019visualbert} proposed a simple single-stream VLP model named VisualBERT. The extracted visual and textual tokens are directly combined and fed into Transformers, where cross-modality fusion can be performed implicitly. Similar to VisualBERT, several concurrent studies such as Unicoder-VL~\citep{li2019unicoder}, VL-BERT~\citep{su2019vl}, and UNITER~\citep{chen2020uniter} also adopt the single-stream architecture. These VLP studies are similar in the following aspects: 1) They all utilize an object detection backbone to compute image embedding. 2) Masked language modeling task is adopted by all of them. 3) They all adopt the single-stream BERT architecture. 
They differ from each other in their pre-training methods and datasets, as shown in Table \ref{table:work comparison}.

\noindent\textbf{Dual Stream Models: } ViLBERT~\citep{lu2019vilbert} and LXMBERT~\citep{tan2019lxmert} are pioneering works to extend BERT to dual-stream VLP models. They are pre-trained on the Conceptual Captions dataset~\citep{sharma2018conceptual} and leverage a pre-trained Faster R-CNN model~\citep{ren2017faster} to detect regions as visual tokens. ViLBERT processes visual and textual tokens separately with two parallel streams which can fuse cross-modality information through cross-attention layers when needed. In other words, ViLBERT assumes the different processing architectures for vision and language. Its cross-modal fusion is designed to be sparse and explicit between the two processing pipelines. LXMBERT differs from ViLBERT by decoupling intra-modal and inter-modal processing. More specifically, visual and textual tokens are encoded separately in the first phase and then fed into a cross-modality encoder to produce the joint representation.

\noindent \textbf{Other Fusion Methods:} Fundamentally, single-stream modeling and dual-stream modeling differ in the fusion time, where single-stream fuses different modalities in an earlier stage while dual-stream prefers to extract high-level features of each modality before fusion. SemVLP~\citep{li2021semvlp} proposed to combine the two prevalent modeling architectures by training them iteratively. Such an approach takes advantage of both architectures and performs cross-modality semantic alignment on both low-level and high-level. Especially, the Transformer encoder is shared between both modeling methods with an additional cross-modal attention module in the dual-stream encoder, which is found to contribute to the semantic alignment and reduce parameters. 

Most VLP models attempt to encode vision and language into separate tokens that interact with each other explicitly or implicitly through modality fusion. Another line of VLP models alternatively attaches visual tokens to textual tokens based on object detection models. B2T2~\citep{alberti2019fusion} proposed to fuse the features of detected objects in textual tokens, based on which MLM and ITM are performed in pre-training. In B2T2, a token $T$ can be expressed as
\begin{equation}
    T=t+\sum_{i=1}^n (h_theta (b_i)+b_i)
\end{equation}
where $t$ is the original textual embedding, $n$ is the number of detected objects whose label is token $t$, $b_i$ is the embedding of the $i$-th object's bounding box, and $h_theta (b_i)$ denotes the extracted visual feature from the bounding box. B2T2 also analyzes the stages of fusing objects and textual tokens. The result indicates the effectiveness of early fusion.
\par
\noindent\textbf{Early Attempts to Bridge Modality Gap: }
To enable both generation and understanding tasks, ~\citet{zhou2019unified} proposed a unified vision-language pre-training approach. It introduces two mask schemes namely bidirectional attention mask and sequence-to-sequence mask to empower understanding and generation tasks, respectively. It is worth noting that this unified VLP approach only adopts MLM during pre-training and achieves a competitive performance on image captioning and VQA. 12-in-1~\citep{lu202012} extended multi-task training to four broad tasks pre-trained on 12 datasets. The experimental results indicate multi-task training can consistently improve the performance of downstream tasks and yield a more lightweight model with fewer parameters.\par

VILLA~\citep{gan2020large} introduced adversarial training at the embedding level of visual and textual tokens based on the design of UNITER~\citep{chen2019uniter}. It performs adversarial training by adding perturbations in the embedding space as regularization and yields decent performance improvement.\par
 
Motivated by the knowledge masking scheme of ERNIE~\citep{sun2019ernie}, structured knowledge is first incorporated in the VLP models in ERNIE-ViL~\citep{yu2020ernie}. {To develop better cross-modality semantic alignments by constructing scene graphs, ERNIE-ViL proposes scene graph prediction tasks to model objects, attributes, and relationships in the graph to learn object-level and attribute-aware representation.} Incorporating knowledge in cross-modality training is challenging and remains an open problem.\par

\noindent\textbf{Grid \& Patch features: } While the prevalence of region feature embedding facilitates the training of VLP models, it also restricts the scalability and generalization capability of VLP models. We can analyze the weakness of region features from Faster R-CNN as follows.
\begin{itemize}
    \item Limited categories: Visual feature is limited by object detection models which are trained on relatively small datasets with predefined object categories. For example, the widely adopted Faster R-CNN model in BUTD~\citep{Anderson_2018_CVPR} is trained on VG with a fixed number of 1594 object classes and 524 attributes.
    \item Low quality: Region features often suffer from low quality as the Faster R-CNN models are trained on small well-labeled datasets ~\citep{anderson2018bottom}.
    \item Lack context: Region features extract RoI features that belong to certain categories without any background information, neglecting semantic relationships between these region features. In reality, these semantic relationships are important.

\end{itemize}
PixelBERT~\citep{huang2020pixel} attempted to break this limitation and fully utilizes visual information by directly learning from pixel features. Instead of utilizing all the pixels as visual features, to reduce computation cost and improve model robustness, a fixed number of 100 pixels are randomly sampled during pre-training. However, the experimental results indicate that random sampling only slightly improve the performance, less than $0.5$ VQA score in downstream tasks. 

SOHO~\citep{huang2021seeing} is another pioneering work that leverages grid features for cross-modality understanding. To learn a semantically comprehensive representation for visual context, SOHO proposes to learn a VD for visual tokenization. VD is learned by first obtaining high-level features from a convolutional network, which are then grouped 
according to feature similarity and fed into a moving-averaged encoder to dynamically update VD. As visual embeddings are trainable, SOHO is an end-to-end pre-training framework that directly learns from pixels, eliminating the need for bounding boxes. With this dynamic VD updating during training, the serial number of each token in the VD can be considered as a label just like language tokens, making it natural to perform masked vision modeling. For pre-training tasks, SOHO proposes a novel MVMVD method (described in \ref{section: maskVisionModelling}) to mask all the visual tokens of the same label simultaneously in an image to avoid any information leakage.

Image embeddings based on regions or grids as aforementioned are computationally heavy and the extracted high-level features prevent early fusion of cross-modality information. Inspired by ViT~\citep{dosovitskiy2020image}, ViLT~\citep{kim2021vilt} adopts simple linear projection of image patches as visual embedding, which greatly accelerates pre-training by $10$ times with competitive results. It implies that instead of designing novel visual embedding, designing better modality-fusion could be the key to improving the representation of VLP models.

\noindent\textbf{Improve Aligned Representation: } 
Vision-language-aligned representation is a fundamental goal in VLP. To achieve this goal, some works propose to adopt additional object-level data in VLP. For example, many VLP methods adopt RoI region features with detection models. However, the detected object tags as an important component are not explicitly modeled in VLP models. To leverage this additional information, Oscar~\citep{li2020oscar} introduced object tags as anchor points to help learn cross-modality-aligned representation. This learning process is empirically natural as the detected object tags often appear in the image-paired text, which helps align vision and language. In addition, training with object tags contributes to learning co-occurrence of objects. Therefore, Oscar yields significant improvement in the downstream understanding and generation tasks. However, the drawback of Oscar is also obvious as it relies on well-labeled image-caption datasets, making it hard to scale. 

As VLP models are limited by inadequate well-aligned (image, caption) pairs, VIVO~\citep{hu2020vivo} proposed to scale up pre-training using a large amount of (image, tag) pairs. VIVO adopts a Hungarian matching loss to perform masked tag prediction, which enables visual vocabulary learning and improves the model generalization ability to describe novel objects in downstream tasks. As a result, it surpassed human performance for the first time on the Novel Object Captioning at Scale (NoCaps)~\citep{agrawal2019nocaps} benchmark. Scaling VLP models based on RoI features also calls for more powerful visual representations. VinVL~\citep{zhang2021vinvl} followed BUTD~\citep{anderson2018bottom} and develops an improved object detection model for VLP with a larger training dataset. More specifically, it adopts ResNeXt152-C4 and merges four public datasets including VG, COCO, Objects365, and OpenImagesV5 for large-scale training. VinVL yields significant improvement on VLP models like VIVO and Oscar and achieves top results on the NoCaps, image captioning, and VQA  leaderboards.\par

\section{Scale up models and data}
Though inspiring progress has been made in the vision-language joint representation, most aforementioned studies primarily focus on object-level representation to pursue good cross-modal alignment. However, they take a strong assumption: image and text pairs are well labeled, which restricts training datasets to relatively small "gold-labeled" datasets. For example, the largest public dataset widely used for VL pre-training is Conceptual Captions~\citep{sharma2018conceptual} with three million image-text pairs. To obtain richer semantics and stronger generalization capability, larger weakly-labeled datasets such as web-crawled datasets are greatly desired. CLIP~\citep{radford2021learning} and DALL-E~\citep{gu2021zeroshot} are the first successful practice to utilize large-scale ($400$M image-text pairs for CLIP and $250$M for DALL-E) web-crawled data for pre-training. Motivated by the success of CLIP and DALL-E, several recent works further built more powerful models with even larger datasets. 

This section aims to introduce models trained with large-scale weakly-labeled datasets.
The section is divided into two parts. The first part includes works utilizing large-scale datasets for visual understanding such as CLIP, ALIGN~\citep{jia2021scaling}, SimVLM~\citep{wang2021simvlm} and Florence~\citep{yuan2021florence}. The second part contains visual generation models empowered by large-scale datasets such as DALL-E, GODIVA~\citep{wu2021godiva} and NÜWA~\citep{wu2021nuwa}. 

\subsection{Visual Understanding}
The core idea of CLIP is the training method. Instead of training to predict masked visual or textual tokens as in other VLP methods, CLIP learns to recognize paired image and text. Given a batch of $N$  (image-text) pairs, the goal is to predict which of the $N\times N$ possible pairs are matched pairs (positive samples) and which are unmatched pairs (negative samples). After pre-training, CLIP can perform zero-shot image classification by using phrases such as "a photo of" plus a category name as prompts to tell the model which categories an input image is the most similar to. Compared with fully supervised baselines, zero-shot CLIP outperforms the baseline on $16$ of $27$ datasets. 

Similar to CLIP, ALIGN~\citep{jia2021scaling} also adopts a dual encoder model with a contrastive loss for zero-shot tasks. It utilizes a larger raw dataset with $1.8B$ image-text pairs. ALIGN outperforms CLIP on many zero-shot visual tasks, which proves that a larger dataset leads to better performance. Except for vision tasks, ALIGN also outperforms previous work on image-text retrieval tasks.
SimVLM~\citep{wang2021simvlm} developed a new approach to VL pre-training. It follows a simple prefix language modeling objective to predict the next token in an autoregressive way. It achieves competitive results on multiple VL tasks and has the ability for text-guided zero-shot learning. Unlike previous works that adopt coarse (image-level) representations and static (image) data, Florence~\citep{yuan2021florence} adopts fine-grained (object-level) representations and extends to dynamic (video) data. For object-level representations, Florence adds an adaptor Dynamic Head~\citep{Dai_2021_ICCV} to the image encoder and trains with an extra object detection dataset. Through pre-training on $900M$ pairs of image-text pairs, Florence achieves new state-of-the-art results in a majority of 44 representative benchmarks.

Apart from zero-shot classification, CLIP can also help detection. For example, ViLD~\citep{gu2021zeroshot} proposes a zero-shot detector via CLIP distillation. Other studies show that CLIP can learn multi-modal features which are more like neurons in the human brain ~\citep{goh2021multimodal} and can help VL tasks \citep{shen2021clip}.

\subsection{Visual Generation}
\label{t2i}
Besides visual understanding, large-scale weakly-labeled image-text-paired data can also assist text-to-image generation.
\citet{ramesh2021zeroshot} developed an image generation system called DALL-E. DALL-E converts images into discrete visual tokens using a discrete variational auto encoder (dVAE) so that a (text, image) pair can be viewed as a single stream of data. During training, the text-image stream is fed into a decoder only Transformer. As for the attention mask, each image token can see all the text tokens. The attention mask among text tokens is standard causal mask. And image-to-image attention use either row, column or convolutional attention mask. In inference time, given text tokens, the generation process is to predict image tokens in an auto-regressive way as in GPT. DALL-E shows impressive results in four aspects: creating anthropomorphized versions of animals and objects, combining unrelated concepts, rendering text, and applying a transformation to existing images.

Inspired by the training method in DALL-E, \citet{wu2021godiva} proposed a method named GODIVA to generate videos from the text. Similar to DALL-E, GODIVA tokenizes each frame of the video and concatenates the text and visual tokens sequentially as a stream to train the model. DALL-E and GODIVA 
are designed for text-to-image generation and text-to-video generation, respectively, while ~\citet{wu2021nuwa} proposed a unified visual generation model which achieves state-of-the-art results on $8$ downstream tasks including text-to-image, text-to-video, video prediction, etc. They proposed a $3$D Transformer that is able to encode all three data formats including text ($1$D), images ($2$D), and videos ($3$D). To better attend on videos, they designed a 3D Nearby Attention to apply attention along both spatial and temporal axes.

\section{Future Trends}
In the last few years, we have witnessed how VLP models scale to use large quantities of weakly-labeled and more diverse data. In the future, the models and data will continue to scale up to achieve stronger modality cooperation and even unified representation. In addition, incorporating knowledge can further empower VLP models to gain better generalization abilities. In this section, we will discuss these future trends.

\subsection{Toward Modality Cooperation}
Besides improving cross-modal tasks with VL datasets, modal cooperation is emerging in pre-training to boost the performance of both single-modal tasks and multi-modal tasks. Modal cooperation is to help different modalities to help each other and learn better representation. For example, improve language tasks with vision data and improve cross-modal tasks with single-modality data.
\subsubsection{Improve Language Tasks with Vision Data }
\
\newline
Improving language learning with visual information has been explored on a broad range of language tasks, 
including machine translation~\citep{ive2019distilling,zhang2019neural}, 
semantic parsing~\citep{shi2019visually,kojima2020learned}, 
and language grounding~\citep{bordes2019incorporating,kiela2018learning}. 
These studies are tailored for specific language tasks and may suffer from modality discrepancies. 
~\citet{tan2020vokenization} proposed a general pre-training model for language representation with a visual aid, in which a "vokenization" model was introduced to extrapolate vision-language-alignment from image captioning datasets to pure-language corpora. More specifically, the "vokenization" model is trained with image-text matching to construct a visual image vocabulary, which is leveraged to map text tokens in language-only datasets to the retrieved images with the highest score. The experimental results show that it can yield additional improvement over self-supervised language models.
\subsubsection{Improve Cross-Modal Tasks with Single-Modality Data.} To address the data shortage issue, some VLP models utilize extra single-modal data to improve the representation capability. For example, in an image-text dataset, text is usually short with several tokens, which restricts the textual representation. Therefore, VL-BERT~\citep{su2019vl} adds additional linguistic corpora to improve the language part in cross-modal tasks.

\subsection{Toward General Unified-Modality}

Thanks to the Transformer architecture, researchers have achieved remarkable progress in both single-modal and multi-modal representation learning. In previous sections, we have discussed multi-modal representation and modal cooperation, which connect vision and language in different ways. A more ambitious goal is to build a general representation model which can unify multiple modalities. As a pioneering work, UNIMO~\citep{li2020unimo} proposed a unified pre-training model, which can handle both single-modal and multi-modal downstream tasks including understanding and generation. Trained with a huge amount of single-modal as well as cross-modal data including BookWiki~\citep{zhu2015aligning} and OpenWebText (language data), OpenImages~\citep{krasin2017openimages} and COCO~\citep{lin2014microsoft} (image data), and COCO \citep{lin2014microsoft}, Visual Genome \citep{krishna2016visual} , Conceptual Captions~\citep{sharma2018conceptual} and SBU~\citep{ordonez2011im2text} (image-text data). As a result, UNIMO improves many single-modal and multi-modal downstream tasks by a large margin. Another interesting work is a general-purpose vision system developed by \citet{gupta2021general} for a series of vision and cross-modal tasks.

\subsection{VL+Knowledge}
Many VL tasks require common sense and factual information beyond training datasets. However, most VLP models do not have a mechanism to consume extra knowledge. ERNIE~\citep{sun2019ernie} proposed a multi-stage knowledge-based masking strategy. Instead of directly adding knowledge embedding, it masks language in three levels, which are basic-level, phrase-level, and entity-level masking. For entity-level masking, the model masks a whole entity rather than a sub-word. Such entities include persons, locations, organizations, products, etc. There are also ways of integrating knowledge into VLP models. \citet{shevchenko2021reasoning} proposed to directly inject knowledge embeddings into a vision-language Transformer. They first build a knowledge base (KB) with knowledge embeddings and then match the sentences in the training data with knowledge embeddings. During training, they use an auxiliary loss to encourage the learned representation to be aligned with knowledge embeddings. Although there have been some works seeking to integrate knowledge into VLP models, there are still many challenges to be addressed such as how to efficiently utilize large wikidata with high noise and how to learn from knowledge in an explainable way. 

{
\bibliographystyle{plainnat}
\bibliography{reference}
}
\end{document}